\documentclass[abstracton]{scrartcl} 
\usepackage{epsf, pstricks, graphics, graphicx} 
\usepackage{multirow} 
\usepackage{endnotes} 
\usepackage[backend=biber, citestyle=numeric, bibstyle=numeric,sortcites]{biblatex}
\usepackage{float} 
\usepackage{marvosym} 
\usepackage{threeparttable}
\usepackage{rotating} 
\usepackage{setspace} 
\usepackage{multicol} 
\usepackage[T1]{fontenc} 
\usepackage[utf8]{inputenc}
\usepackage{amsmath, amsthm}
\usepackage{amssymb}
\usepackage{mathcomp}
\usepackage{lscape}
\usepackage{multicol}
\usepackage{framed}
\usepackage[german,english]{babel}
\newcommand{\x}[1]{\textit{#1}}

\newtheorem{defn}{Definition}

\begin{document} 
\title{Why machines will \x{still} not rule the world}
\author{Jobst Landgrebe and Barry Smith}
\date{\today}
\publishers{jobst.landgrebe@usi.ch, phismith@buffalo.edu}
\selectlanguage{english}
\maketitle

\selectlanguage{english}


\begin{abstract}
In our book \x{Why machines will never rule the world} \cite{landgrebesmith:2022,landgrebesmith:2025} we argue that artificial general intelligence is mathematically impossible. This is because the human beings and the processes which exhibit intelligence are complex systems whose behaviour cannot be captured by the kinds of models that we can generate with or without computers. Proponents of contemporary machine intelligence respond with two lines of argument: a theoretical one, grounded in the universal approximation theorems for neural networks and the Church-Turing-Deutsch principle; and an empirical one, grounded in rapidly rising scores on standardized benchmarks. In this communication we examine and reject both responses. First, we show serious issues in the physicalist counter-argument based on the Church-Turing-Deutsch principle. Second, we review recent evidence to the effect that prominent benchmarks are compromised by training-data contamination, flawed test construction, and strategic optimization. Our central argument remains: That models required to perform cognitive behaviour in open-ended, thermodynamically complex and non-ergodic environments are not and will not become achievable.
\end{abstract}

\noindent \small Keywords: artificial general intelligence, complex systems, ergodicity, Church-Turing-Deutsch principle, benchmark contamination, Goodhart's law. \normalsize

\section{Introduction}

Few questions in contemporary philosophy of technology are as consequential, or as contested, as whether machines can come to match the general intelligence of human beings. Public discourse on the question has been shaped largely by empirical milestones: each new generation of large language models (LLMs) is introduced alongside a table of benchmark scores, and the steady upward movement of those scores is frequently interpreted as evidence that general intelligence is approaching. This leads to an extreme hyperboly according to which mankind is currently building the tools of its own destruction, which will occur when evil, hyperintelligent AI systems will `switch off' mankind.
In this cultural context we advance in our monograph \cite{landgrebesmith:2022, landgrebesmith:2025} the counter-claim that artificial general intelligence (AGI) is not merely distant or difficult, but that it is impossible, because the systems that would need to be modeled to produce it are of a kind that mathematics cannot adequately model. 

This essay evaluates this thesis against its two sets of main objections. The first is a theoretical counter-argument holding that, because brains are physical systems and physical systems are computable, anything a brain does can in principle be done by a computer. This argument draws on the universal approximation theorems for neural networks \cite{cybenko:1989, hornik:1989} and, in its strongest form, on the Church-Turing-Deutsch principle \cite{deutsch:1985}. The second is an empirical counter-argument holding that benchmark performance demonstrates the progressive acquisition of general capabilities. We argue here that this theoretical argument against our impossibility claim establishes nothing and that recent methodological audits seriously undermine the empirical counter-argument. 

\section{Complex systems limit mathematical modeling}

We build our argument on a distinction between what we call \x{logic systems}\footnote{The familiar `simple' vs. `complex' opposition is unfortunately often confused in the AI literature with the  opposition between what is `simple' and what is `complicated'. Not every complicated system is a \x{complex system} in the technical sense deployed here and articulated in detail in our book.} on the one hand -- by which we mean systems whose behavior can be captured by explicit equations -- and complex systems, which means systems which exhibit certain thermodynamical properties, including evolutionary character, dependent interactions, force overlay, non-equilibrium dynamics, context dependence, and the absence of a stable phase space. Human intelligence, we contend, is the product of precisely such a system -- the human organism embedded in a natural  and social environment -- and its most important capacities, including dialogue and the open-ended spontaneous adaptation to novel situations, are expressions of that complexity. Because the mathematics available to science can model complex systems only partially, approximately, and for limited purposes, we argue that no computer program can reproduce the behavior those systems generate. For this is a behaviour that requires the functioning of the entire system processes, as we can see from neurological or psychiatric diseases, where partial dysfunctions as experienced, for example in intoxication with psychoactive substances, are already detrimental to the brain's functioning.

The argument is not a claim about computational speed or hardware. It is a claim about the availability of models. Computers execute algorithms, and algorithms implement mathematical models; if adequate models of a given system cannot be constructed, then no amount of computation will reproduce that system's behavior. In an earlier paper, we applied a version of this reasoning to natural language, arguing that dialogues create contexts and ever-shifting intentions of dialogue participants that AI models cannot emulate \cite{landgrebesmith:2021}. The formal core of the argument concerns the statistical properties of the processes that complex systems generate, and in particular their lack of ergodic behaviour.
      
\section{Ergodicity and non-ergodicity}
The concept of ergodicity provides the clearest formal statement of why learning from past data may fail to support prediction of future behavior. The following definitions follow the standard measure-theoretic presentation formulated by \textcite{birkhoff:1931}.

\begin{defn}
Measure-preserving dynamical system. A measure-preserving dynamical system is a quadruple $\{\Omega, \mathcal{F}, \mu, T\}$, where $\{\Omega, \mathcal{F}, \mu\}$ is a probability space and $T: \Omega \to \Omega $ is a measurable transformation satisfying $\mu(T^-(A))= \mu(A), \forall A \in \mathcal{F},$ i.e. all $A$ are elements of the $\sigma$-algebra $\mathcal{F}$.
\end{defn}

 \begin{defn} Ergodicity. A system is ergodic if every invariant set is trivial; that is, whenever $T^-(A) = A \rightarrow \mu(A) \in \{0, 1\}$.  \end{defn}

Intuitively, an ergodic system cannot be decomposed into separate regions that its trajectories never leave. Birkhoff's (1931) pointwise ergodic theorem states the operational consequence of this definition. For an ergodic system, an ergodic mapping $T$ and any integrable observable $f$, we have:
	\begin{equation}
\lim_{n\to\infty}\; \frac{1}{n} \sum_{k=0}^{n-1} f(T^k x)=E[f],
	\end{equation}
where $E$ is the expectation of invariant sets of $T.$

For a stochastic process $X(t)$ in continuous time, the corresponding statement is that the time average along a single realization converges to the ensemble average $E[X]$:
	\begin{equation}\label{e:er}
	\lim_{T\to\infty} \frac{1}{T} \int_0^T X(t) dt = E[X].
	\end{equation}

\begin{defn}
Non-ergodicity. A process is non-ergodic with respect to an observable if equation \ref{e:er} is not true -- either because the time-average limit does not exist, or because it exists but differs from $E[X]$ on a set with a positive measure, so that what is observed along one trajectory depends on initial conditions and its past.
\end{defn}

An everyday example is an idealised bottle of water at rest in perfect thermodynamic equilibrium. The particles in the bottle move in an ergodic manner, which means that each will at some point in time return to exactly the same position at which it was at a given moment of (idealised) observation. Or, in other words, the likelihood of every particle being anywhere in the bottle is identical for all molecules over time. If the bottle is heated, put under pressure, shaken or energy is introduced into it in any other manner, the particles behave in a non-ergodic fashion and the particles' behaviour becomes non-ergodic.

\textcite{peters:2019} offers a simple example from economics. Consider a repeated gamble in which wealth is multiplied by 1.5 or by 0.6 with equal probability. The ensemble average of the per-round growth factor is $E[r] = \frac{1}{2}\cdot1.5 + \frac{1}{2}\cdot0.6 = 1.05$,
suggesting growth of 5\% per round. The time-average growth factor experienced by any single individual, however, is the geometric mean:
	$\bar{g} = (1.5 \cdot 0.6)^{\frac{1}{2}} = \sqrt{0.9} \approx 0.949,$
implying a decline of about 5\% per round. Almost every individual trajectory therefore decays toward zero even though the expectation across the ensemble grows. The two averages diverge because the process is multiplicative and therefore non-ergodic in wealth. That said, the fact that the ensemble average is ergodic, while the individual trajectory is not in this simple example does not mean that one can model non-ergodic phenomena as regular; the simple example does not generalise to real complex systems. Regularity here means the presence of recurrent patterns.

\subsection{Machine learning}
The relevance to machine learning follows from the structure of statistical learning theory. Supervised learning, the tool that generates all LLMs and other stochastic models by optimisation, minimizes the empirical risk
	\begin{equation}\label{e:c}
	\hat{R}_n(f) = \frac{1}{n} \sum_{i=1}^n \ell(f(x_i), y_i),	
	\end{equation}

\noindent where $\ell$ is a loss function, $f$ is a machine learning functional and $R$ is risk. $\hat{R}_n(f)$ is used as an estimate of the true risk 
\begin{equation}R(f) = E_{x,y} \sim \mathcal{D}_\kappa[\ell(f(x),y)],\end{equation} where $\mathcal{D}_\kappa$ is a training and test distribution. 

In this setting the ability of the functional to generalise depends on the assumption that training, test and deployment data are drawn from the same fixed distribution $\mathcal{D}_\kappa$. If the process generating the data is non-ergodic -- so that the deployment distribution $\mathcal{D}_1$ is not identical with the training distribution $\mathcal{D}_0$:

\begin{equation}\label{e:a}\mathcal{D}_1 \cup \mathcal{D}_0 \neq \mathcal{D}_0,\end{equation} then even a small empirical risk (reflecting excellent training and test performance) provides no guarantee as to future performance. We argue \cite{landgrebesmith:2022} that this is the necessary condition of human intelligence and the environment in which it operates, since both are complex systems and non-ergodic. It is therefore impossible to sample from a distribution generated by human intelligence adequately, and equation (\ref{e:a}) will always hold.

\section{The Connectionist Counter-Thesis}
\subsection{Universal Function Approximation}
The first theoretical response to our argument appeals to the expressive power of neural networks (NNs). \textcite{cybenko:1989} and \textcite{hornik:1989} proved that feedforward networks with a single hidden layer are universal approximators. This result is often formulated as follows, where it is important to note that a non-polynomial functional $\sigma$ is required to approximate an arbitrary functional using a neural network:

Let $C(X, \mathbb{R}^m)$ denote the set of continuous functions from a subset $X $ of a Euclidean $\mathbb{R}^n$ space to a Euclidean space $\mathbb{R}^m$. Let $\sigma \in C(\mathbb{R}, \mathbb{R})$.

Then $\sigma$ is not polynomial \x{if and only if} for every $n \in \mathbb{N}$, $m \in \mathbb{N}$, compact subspace $K \subseteq \mathbb{R}^n$, $f \in C(K, \mathbb{R}^m), \varepsilon > 0$, there exist $k \in \mathbb{N}$, $A \in \mathbb{R}^{k \times n}$, $b \in \mathbb{R}^k$, $C \in \mathbb{R}^{m \times k}$ such that
\begin{equation}
\sup_{x \in K} \| f(x) - g(x) \| < \varepsilon
\text{~where~} 
g(x) = C \cdot ( \sigma \circ (A \cdot x + b) )
\end{equation}

NN-connectionists infer that, if intelligent behavior is a function from inputs to outputs, then a sufficiently large network can represent it. The theorem, however, carries three restrictions that matter for the present debate. (i) It is an \x{existence result} and says nothing about whether gradient-based training on finite data will ever \x{find} the needed approximating parameters. (ii) It applies to fixed functions on compact domains, whereas complex systems -- such as inanimate and animate entities in nature -- are characterized by open, shifting domains. And (iii) it presupposes further that the target function exists as a stable mapping, which is exactly what non-ergodicity calls into question. The universal approximation theorem therefore does indeed show that representational capacity is not the bottleneck; but it fails to show that the right function can be learned or even specified when it comes to modelling complex systems. It cannot be used to rebut the lack of a regular distribution resulting from non-ergodic processes.
      
\subsection{The Church-Turing-Deutsch Principle}
A deeper response grounds the possibility of AGI in physics rather than in network architecture. The classical Church-Turing thesis, formulated independently by \textcite{church:1937} and \textcite{turing:1936}, concerns effective computability: any function computable by an idealized human following a mechanical procedure is computable by a Turing machine. The thesis is a claim about mathematical procedures that can be realised by idealised machines. \textcite{deutsch:1985} proposed a physical strengthening -- now commonly called the Church-Turing-Deutsch (CTD) principle -- whereby every finitely realizable physical system can be perfectly simulated by a universal computing machine operating by finite means. Because Deutsch showed that classical Turing machines cannot satisfy this principle for quantum systems in the strict sense he intended, he introduced the universal quantum computer as the machine that can.

The physicalist argument against our thesis then proceeds in three steps. 

\begin{enumerate}	  
      \item The human brain is a finitely realizable physical system. 
      \item Its time evolution is governed by physical laws; in the quantum-mechanical description, a closed system with Hamiltonian $H$ evolves under the unitary operator
	  \begin{equation} \label{e:b}
	U(t) = e^{\frac{-iHt}{\hbar}}.	
	  \end{equation}
      \item By the CTD principle, a universal quantum computer can simulate this evolution to arbitrary precision. 
If the brain is a physical system of this kind, the conclusion follows that there is no in-principle barrier to its simulation.
\end{enumerate}	  

This syllogism is however false and establishes nothing against our thesis. First, the CTD principle, though it is widely accepted among physicalists, is a conjecture about physics rather than a proven theorem. More specifically, we show in chapter~7 of \cite{landgrebesmith:2025} that the CTD-thesis is not supported by modern physics, which provides us with only very incomplete models of nature suitable, with small exceptions, only to artificial or highly idealised settings.
Second, the argument establishes simulability given the Hamiltonian and the initial state. We do not claim that a complex system's dynamics are uncomputable once known, but rather that the relevant models, equations, parameters, and initial conditions, cannot be obtained for complex systems. Of course we acknowledge that there are partial models for complex systems -- for example LLMs -- that are prefectly computable, such as the models for mostly\footnote{\textcite{murphy:2025} show that important fundamental principles of syntax are not modelled by LLM.} correct syntax. But the total behaviour of a complex system, such as the cognitive abilities of the human brain, cannot be modelled in this way. Parameterising equation (\ref{e:b}) for a human brain would require specifying $H$ and the initial quantum state of roughly $10^{26}$ interacting particles, together with the environment with which the brain continuously exchanges energy and information. We do not have such models, let alone the ability to measure the initial conditions. Third, even an exact simulation of one brain would not constitute a predictive model of an open environment; it would reproduce one trajectory of a non-ergodic system, inheriting rather than resolving the forecasting problem described in equation (\ref{e:c}).

The CTD principle therefore has nothing to say against our thesis. It claims to remove the suggestion that intelligence involves something non-physical. But we never make this suggestion. Rather, we defend a position we call `nomological monism', which is a version of strict materialism \cite[chapter~2]{landgrebesmith:2025}. All we say is that processes which cannot be mathematically modelled are not computable and that what we introspectively experience as the human mind cannot be modelled.  The AGI hyperbole proponents claim nevertheless that such models can be acquired, and since they cannot support this using theoretical arguments, they turn from theory to what they see as empirical evidence.

\section{So-called benchmark evidence}
The empirical counter-thesis to our claim as to the limits of AI holds that, whatever the theoretical arguments, LLMs have demonstrably acquired general capabilities, as shown by their performance on benchmarks designed to test reasoning, expert knowledge, and open-ended problem solving. The force of this argument depends entirely on the validity of the benchmarks, and recent audits give substantial reason to doubt that validity.

\subsection{Contamination and non-validity of benchmarks}

The claim is regularly made that benchmark performance proves the cognitive abilities of models. An important benchmark to measure the quality of LLMs in software engineering, SWE-bench, is a model evaluation framework consisting of 2294 software engineering problems drawn from real GitHub issues. Its aim is to evaluate whether a model can resolve real issues drawn from open-source Python repositories on GitHub, where correctness is judged by the repositories' own test suites \cite{jimenez:2024}. Its human-validated subset, SWE-bench verified, became a standard metric in frontier model releases, with top systems reporting scores near 80\%. 

But in February 2026, \textcite{openAI:2026} announced that it would stop reporting SWE-bench verified results. Its evaluations team audited 138 problems that a strong model had failed to solve consistently and found that 59.4\% contained material flaws in test design or problem descriptions -- including  tests that rejected functionally correct solutions. The audit also found that frontier models from several developers could reproduce some reference solutions without search (merely using the models' configurations), a strong indication of training-data contamination arising from the public availability of the benchmark and its source repositories. 

To limit contamination, SWE-bench pro, was constructed with held-out, publicly non available commercially sourced codebases. This yielded scores of roughly 23\% for leading models at its release \cite{deng:2025}, compared with scores above 70\% for comparable models on SWE-bench verified. The comparison must be interpreted cautiously, because SWE-bench pro was deliberately designed to contain longer, harder tasks, and some of the gap reflects difficulty rather than contamination. Nevertheless, this finding shows that a score widely presented as evidence of autonomous software engineering capability was, to a material degree, a measure of training configuration and of the idiosyncrasies of a test harness.
Thus the contrast between SWE-bench type model performance under contamination conditions, on the one hand, and contamination-resistant evaluation, where the model is unable to make usage of training-distribution derived configuration, on the other, is striking. 

The practical consequences of this inability to cope with novel situations and with context diversity become evident when one looks at the domain in which LLM-generated output has been used for production purposes, namely software engineering. There the error rates are so high and the inability to take into account the architectural context in AI-generated code increases maintenance costs or leads all too often to outages once the model-generated code is deployed.\footnote{The March 2026 Amazon outages provide a good example: https://ai-analytics.wharton.upenn.edu/wharton-accountable-ai-lab/governing-ai-agents-what-the-amazon-outage-reveals-about-enterprise-risk/ Note that LLM in software engineering are very effective and efficient if their output is carefully reviewed.} 

\subsubsection{Knowledge Benchmarks and Web-Scale Contamination}

Similar concerns apply to knowledge-intensive benchmarks such as MMLU-Pro \cite{wang:2024} and GPQA \cite{rein:2024}, which are often cited as evidence of expert level reasoning. Because LLMs are trained on web-scale corpora, benchmark items and their answers can enter training data through many indirect routes. \textcite{deng:2024} showed that some commercial models could guess deliberately masked incorrect options in multiple-choice test sets, a behavior difficult to explain without assuming exposure to the data. A systematic review of 55 studies \cite{nourbakhsh:2026} proposed a taxonomy of contamination ranging from exact duplication to task-level leakage and concluded that no existing detection method is consistently reliable across contamination types, model-access settings, and training stages. The implication is not that every high score is contaminated, but that contamination can rarely be ruled out, which weakens the inference from score to capability.

A related problem arises when models are evaluated by other models. LLM judges have been shown to favor longer and more verbose responses \cite{zheng:2023} and to be susceptible to authority cues and superficial features of presentation such as citations and formatting, independent of substantive correctness \cite{chen:2024}. To the extent that evaluations reward the surface features of correct-looking answers, what is being measured is fluency in the conventions of expert discourse rather than the reasoning those conventions normally signal.

In other words, what we see when LLMs perform well at benchmarks is the result of a configuration with training data that contain many of the data which are used to score the models. In the sense of equations (\ref{e:c})-(\ref{e:a}) this means that $\mathcal{D}_0 \cap \mathcal{D}_1 \neq \varnothing$ with the model performance depending on the size of the intersection between deployment ($\mathcal{D}_0$) and training ($\mathcal{D}_1$) set.

\subsection{Human-preference leaderboards}

Chatbot Arena is an LLM rating scheme allowing anyone to submit a prompt and subsequently rank two anonymous responses from different models.
It was designed to address the static benchmark problem by collecting large numbers of pairwise human preference judgments on open-ended prompts \cite{chiang:2024}. \textcite{singh:2025}, however, documented systematic distortions in its leaderboard. Among their findings was that some providers privately tested many model variants before public release and published only the best-performing one. They also found that \begin{quote}proprietary closed models are sampled at higher rates [$\dots$] and have fewer models removed from the arena than open-weight and open-source alternatives. Both these policies lead to large data access asymmetries over time. $[\dots]$ Together, these dynamics result in overfitting to Arena-specific dynamics rather than general model quality.\end{quote}  Taken together it seems that preference leaderboards measure the persuasiveness of outputs to typical users on typical queries rather than performance in sustained, consequential, multi-turn tasks.

\subsection{Goodhart's law} 
These failures share a common structure, captured by Goodhart's law \cite{goodhart:1984}. He observed that statistical regularities in monetary policy tended to collapse once they were used for control. \textcite{strathern:1997} generalized the point into its familiar form, that a measure which becomes a target ceases to be a good measure. A benchmark is valid only so long as performance on it is correlated with the underlying capability it was designed to detect. Once the benchmark becomes the object of optimization -- through training on public data that includes its items, through selective reporting, or through tuning toward the preferences of its raters -- that correlation weakens, and the score increasingly measures the optimization process itself.

This analysis is directly related to the formal argument of the preceding sections. A static benchmark is a fixed sample from a fixed distribution. High performance on such a benchmark shows that a model has achieved low empirical risk with respect to that distribution (equation (\ref{e:c})). It cannot, by construction, show that the model will perform well when the distribution shifts (equation (\ref{e:a})). Benchmark contamination is in this sense an extreme case of the general problem: when the test distribution has leaked into the training distribution, the evaluation measures interpolation within known data rather than adaptation to the novel, non-ergodic conditions that we know are an essential characteristic of real environments.

\section{Discussion}
It would of course be a mistake to conclude that the benchmark evidence vindicates our impossibility thesis. Our claim is modal and universal: no machine will ever achieve artificial general intelligence (AGI). Benchmark failures are contingent and particular: specific measurement instruments have been compromised in specific ways. The fact that SWE-bench verified overstated capability does not show that capability is absent, only that this instrument cannot establish its presence. Moreover, models continue to show over time measurable, if more modest, performance on contamination-resistant evaluations. Continuously refreshed benchmarks that draw tasks from after a model's training cutoff offer partial remedies to the contamination problem. The appropriate conclusion is therefore asymmetric: The audits substantially weaken the empirical case for general intelligence, but they do not by themselves supply an empirical case against its possibility.

Conversely, proponents of the connectionist position cannot rely on the CTD principle to make the case for AGI. In-principle simulability is compatible with what we assert as the in-practice unobtainability of relevant models. Can learning systems acquire models that remain reliable under sustained distributional shift? Can evaluation methods be designed that detect such reliability without themselves becoming targets of optimization? We remain confident that the answer to both of these questions is negative.
      
We argue that AGI is impossible because complex systems -- which means all organisms and also all of inanimate nature -- generate non-ergodic behavior, behavior  that mathematical models, and therefore computers, cannot capture. We have provided a formal account of the ergodicity that lies at the heart of this argument and shown why non-ergodicity undermines the statistical assumptions on which machine learning's generalization guarantees rest. We have reconstructed the strongest theoretical arguments, grounded in the universal approximation theorems and the Church-Turing-Deutsch principle, and shown that these results establish representational and in-principle computational adequacy without giving any evidence that the requisite models could ever be obtained. Contrarywise, both approaches make it clear why intelligence cannot be modelled, since both show that complex system behaviour is out of scope for mathematics.
      
Importantly, though it is impossible to model complex non-ergodic processes \x{mathematically}, the human brain can react to non-ergodic processes generating responses adequate to the situation without any preparation. That is what we call intelligence \cite[secti.~3.2]{landgrebesmith:2025}.

We also reviewed evidence that leading benchmarks have been compromised by contamination, flawed construction, and strategic optimization, in conformity with Goodhart's law. The benchmarks do not show that machines have mastered complexity; they show, in large part, that developers have become adept at optimizing for benchmarks. The failure of a measurement instrument is of course not proof of the impossibility of what it was meant to measure. On the other hand, however, we postulate that evaluation regimes that are built to challenge algorithms with non-ergodic data that are held out at training time and continuously renewed and which are resistant to optimization will provide the proponents of AGI with evidence to the effect that machines will not become intelligent since they will not be able to cope with token sequences that are absent from their training material. This is already obvious from challenges to so-called `reasoning models' presented in \cite{shojaee:2025}. 

We do not think that the human mind is capable of creating mathematics that can holistically model complex systems, neither on its own with explicit modelling or with the help of optimisation algorithms developed by humans that yield huge implicit models such as LLMs or multi-modal models. Furthermore, the new edition of our book contains material regarding practical intelligence \cite{smith:2024}, a capability which operates below the level of consciousness which yet dominates human behaviour and grounds cognitive intelligence. This is intelligence of a form that is not even accessible to machine configuration, since no pertinent training data could ever be made available.

\printbibliography
\end{document}